\documentclass[sigconf]{acmart}
\usepackage{booktabs} % For formal tables
\usepackage{etoolbox}

\setcopyright{none}
\acmConference[SysML 2018]{SysML}{Stanford, California USA}{}
\acmYear{February 2018}
\acmDOI{}
\acmISBN{}
\graphicspath{{figs/}}

\begin{document}
%\title{$\mathbf{CascadeC_{N_N}:}$ A toolflow for... }
\vspace{-0.5cm}
\title{ $\mathbf{Cascade^{C_{N_N}}}$: Pushing the performance limits of quantisation}
\vspace{-0.5cm}
%\title{$\mathbf{Cascade^CN_N:}$ A toolflow for... }

%\titlenote  %{ Produces the permission block, and
%  copyright information}
%\subtitle{Extended Abstract}

\author{Alexandros Kouris}
%\orcid{1234-5678-9012}
\affiliation{%
  \institution{Dept. of Electrical and Electronic Eng. \\Imperial College London}
  %\streetaddress{P.O. Box 1212}
  %\city{Dublin}
  %\state{Ohio}
  %\postcode{}
}
\email{a.kouris16@ic.ac.uk}

\author{Stylianos I. Venieris}
%\orcid{1234-5678-9012}
\affiliation{%
  \institution{Dept. of Electrical and Electronic Eng. \\Imperial College London}
  %\streetaddress{P.O. Box 1212}
  %\city{Dublin}
  %\state{Ohio}
  %\postcode{}
}
\email{stylianos.venieris10@ic.ac.uk}

\author{Christos-Savvas Bouganis}
%\orcid{1234-5678-9012}
\affiliation{%
  \institution{Dept. of Electrical and Electronic Eng. \\Imperial College London}
  %\streetaddress{P.O. Box 1212}
  %\city{Dublin}
  %\state{Ohio}
  %\postcode{}
}
\email{christos-savvas.bouganis@ic.ac.uk}

% The default list of authors is too long for headers.
%\renewcommand{\shortauthors}{B. Trovato et al.}

\begin{abstract}
\vspace{-0.1cm}
This work presents \textit{CascadeCNN}, an automated toolflow that pushes the quantisation limits of any given CNN model, to perform high-throughput inference by exploiting the computation time-accuracy trade-off. Without the need for retraining, a two-stage architecture tailored for any given FPGA device is generated, consisting of a low- and a high-precision unit. A confidence evaluation unit is employed between them to identify misclassified cases at run time and forward them to the high-precision unit or terminate computation. Experiments demonstrate that \textit{CascadeCNN} achieves a performance boost of up to 55\% for VGG-16 and 48\% for AlexNet over the baseline design for the same resource budget and accuracy.
\end{abstract}

\maketitle

\vspace{-0.2cm}
\section{Introduction}
\vspace{-0.1cm}
% While Convolutional Neural Networks are becoming the state-of-the-art algorithm in various Machine Vision tasks such as image classification \cite{krizhevsky2012imagenet}, object detection \cite{redmon2016you} and semantic segmentation \cite{badrinarayanan2015segnet}, they are challenged to deal with problems of continuously increasing complexity. The significant advancement of CNNs came with: increased number of layers \cite{szegedy2015going}, increased number of kernels \cite{zeiler2014visualizing}, decreased stride size \cite{Simonyan14c} and more complex architectures combining multiple CNN layers \cite{he2016deep}. Thus, significant complexity costs are introduced by this advancement in terms of both computation requirements and memory footprint. It is crucial, that the high computation and memory requirements of such methods are alleviated to allow their application on real-world tasks, dealing with vast amounts of data. By exploiting CNN model redundancy, efficient implementations of CNNs take advantage of the reduced model size offered by numerous recently studied compression techniques \cite{hubara2016quantized} \cite{han2015deep} \cite{lin2016fixed} \cite{wu2016quantized}, used to reduce the computation payload and allow their efficient deployment on the available processing platforms (GPUs, FPGAs etc). 

While Convolutional Neural Networks are becoming the state-of-the-art algorithm in various Machine Vision tasks \cite{krizhevsky2012imagenet}\cite{redmon2016you}\cite{badrinarayanan2015segnet}, they are challenged to deal with problems of continuously increasing complexity. The significant advances of CNNs came with increased number of layers \cite{simonyan2014very}, increased number of kernels \cite{zeiler2014visualizing} and more complex architectures \cite{szegedy2015going}\cite{he2016deep}, which introduce substantial %complexity 
costs in terms of computational and memory resources. %requirements. 
To deploy CNNs in real-world tasks which deal with vast amounts of data, it is necessary that the high computation and memory requirements of such models are alleviated. %By exploiting the redundancy in CNN models, efficient implementations of CNNs can take advantage of a reduced model size, as demonstrated by numerous recently proposed compression techniques \cite{hubara2016quantized} \cite{han2015deep} \cite{lin2016fixed} \cite{wu2016quantized}, to enable the efficient deployment on processing platforms. %, used to reduce the computation payload and allow their efficient deployment on the available processing platforms (GPUs, FPGAs etc). 
To this end, numerous compression and precision quantisation techniques \cite{hubara2016quantized}\cite{han2015deep}\cite{lin2016fixed}\cite{wu2016quantized} have been proposed which exploit the redundancy in CNN models to enable the efficient deployment of CNNs on processing platforms.

In this context, FPGAs constitute a promising platform for CNN inference due to their customisability which enables the use of optimised low-precision arithmetic units to achieve high performance at a low power envelope \cite{Venieris_2017c}. Existing FPGA-based CNN accelerators have produced hardware designs that span from uniform 16-bit activations and weights \cite{Venieris_2016}\cite{Yufei_Ma_2017b} % and mixed 16-bit activations and 8-bit weights \cite{Yufei_Ma_2017}
with minimal effect on accuracy, down to very high-performance binarised networks \cite{Umuroglu_2017} but with a significant accuracy loss. In this setting, given a fixed resource budget, the attainable performance for a given error tolerance is limited by the shortest wordlength that meets the error bound.

In this paper, we propose \textit{CascadeCNN}, a novel automated approach of pushing the performance of precision-quantised CNN models under the same resource budget, with negligible accuracy loss. \textit{CascadeCNN} employs a low-precision processing unit to obtain rapid classification predictions together with a parametrised mechanism for identifying misclassified cases based on prediction confidence. Such detected cases are recomputed on a high-precision unit to restore application-level accuracy and meet user-specified limits. \textit{CascadeCNN} considers the error tolerance and the target CNN-device pair to select quantisation scheme, configure the confidence evaluation mechanism and generate the cascaded low- and high-precision processing units. 

%the precision quantisation of a given CNN model below the limits of acceptable accuracy loss in order to gain performance on the inference phase, combined with a mechanism for identifying misclassified cases based on prediction confidence. Such detected cases are recomputed on a higher precision unit to restore application-level accuracy and meet user-specified limits, without requiring data for retraining the CNN model. The system is deployed on an FPGA-based two-stage hardware architecture, optimised for the particular CNN model and FPGA device, based on performance modelling.
 
 \section{Cascade CNN}
 \vspace{-0.05cm}
 \subsection{Overview}
      \vspace{-0.1cm}
%  In more detail, \textit{CascadeCNN} is an automated toolflow which given a high-level CNN description (i.e. Caffe model), the available resources of any target FPGA platform (LUTs, DSPs, memory BW etc) and an application-level error tolerance (top1/top5 classification error for example) along with a small application evaluation set provides a two-stage tailored hardware architecture, optimised for the particular CNN model and FPGA device, consisting of:
%  \begin{itemize}
%      \item A Low-precision Unit for high-throughput CNN inference with degraded application-level accuracy
%      \item A tunable Confidence Evaluation Unit to detect wrongly classified samples caused by extreme model quantisation
%      \item A High-precision unit to re-process captured misclassified samples for restoring accuracy to meet the desired error threshold
%  \end{itemize}
%  Providing that this work targets throughput demanding applications, each unit is mapped across the whole FPGA device, with an intermediate reconfiguration step to switch between the low- and high-precision stages, the cost of which is alleviated via batch processing.

\begin{figure}
	\centering
	 	    \vspace{0.3cm}
	\includegraphics[trim =0mm 0mm 0mm 0mm, width=0.90\columnwidth]{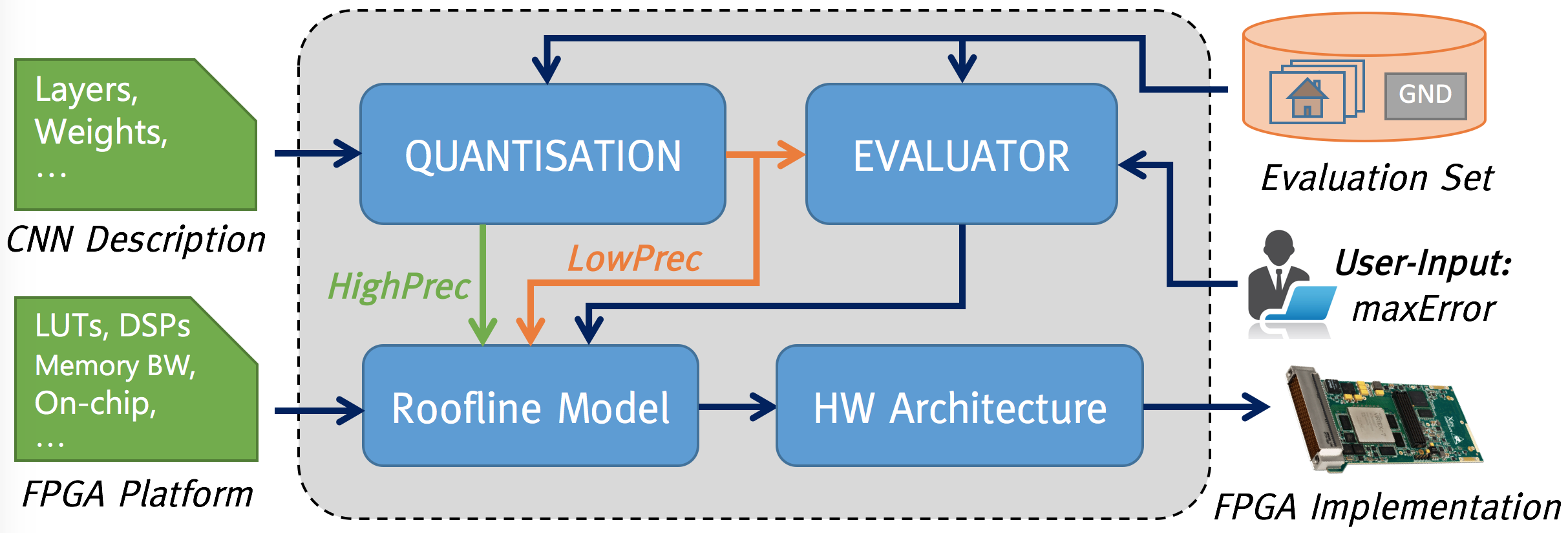}
		    \vspace{-0.35cm}
	\caption{High-level \textit{CascadeCNN} toolflow}
	\label{fig:toolflow}
	    \vspace{-0.22cm}
\end{figure}

\begin{figure}[h] 
 		    \vspace{-0.15cm}
	\centering
	\includegraphics[trim =0mm 0mm 0mm 0mm, width=0.85\columnwidth]{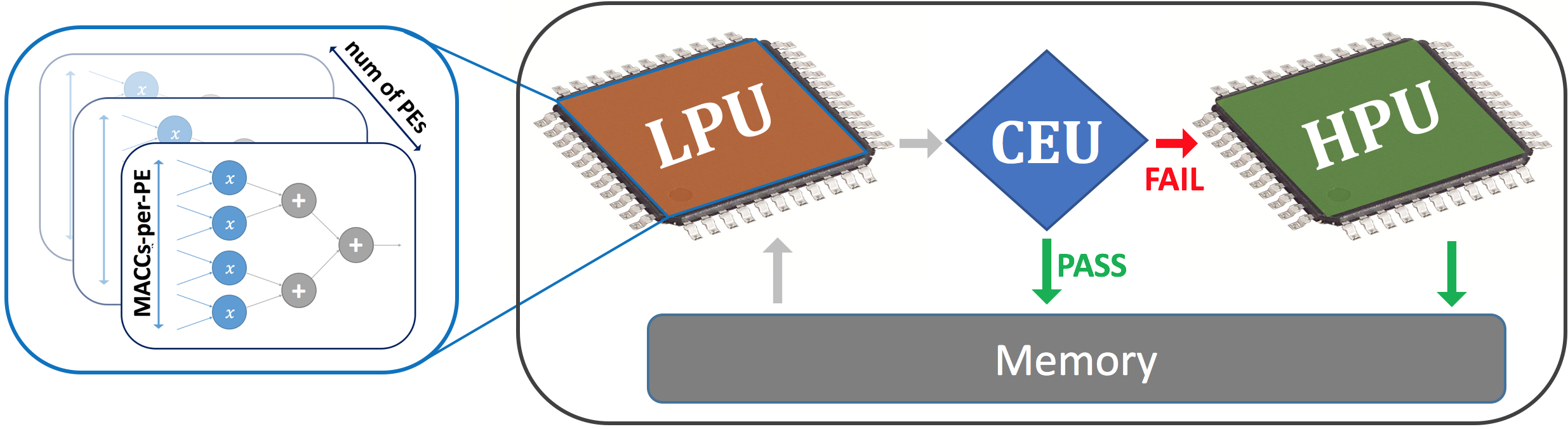}
			    \vspace{-0.3cm}
	\caption{\textit{CascadeCNN} architecture }
	\label{fig:arch}
		    \vspace{-0.55cm}
\end{figure}

Fig. \ref{fig:toolflow} shows the processing flow of \textit{CascadeCNN}. The framework is supplied with a high-level description of a trained CNN model (i.e. Caffe model), the available computational and memory resources of the target platform and an application-level error tolerance in a user-defined metric (e.g. top-1/top-5 classification error), along with a small evaluation set.
%\textit{CascadeCNN} is an automated toolflow that exploits the computation time-accuracy trade-off to accelerate CNN inference. \textit{CascadeCNN} is supplied with a high-level description of a trained CNN model (i.e. Caffe model), the available computational and memory resources of any target FPGA platform and an application-level error tolerance in a user-defined metric (e.g. top1/top5 classification error), along with a small evaluation set.
\textit{CascadeCNN} searches the architectural design space and generates a two-stage hardware architecture, optimised for the particular CNN model and target device. The generated system (Fig. \ref{fig:arch}) consists of:
      \vspace{-0.1cm}
\begin{itemize}
 \item A low-precision unit (LPU) which employs low-precision arithmetic to trade lower accuracy with high-throughput CNN inference.
 \item A high-precision unit (HPU) which guarantees the same accuracy level as the reference model.
 \item A tunable Confidence Evaluation Unit (CEU) that detects samples that were wrongly classified by the LPU and redirects them to HPU for re-processing.
\end{itemize}
      \vspace{-0.1cm}
The key idea behind the proposed approach is that during the execution of the system, the LPU will process the whole workload, while the HPU will only process a fraction of it, based on the CEU's evaluation of classification confidence on LPU's predictions, reducing its memory and compute requirements. Moreover, the accuracy loss that is induced due to the extreme model quantisation of the LPU is restored to meet the user-specified error threshold.

%At run time, all input samples are initially processed in batches by the LPU. The CEU examines the outputs of the LPU and determines the classification confidence for each sample. Only the samples that are detected as wrongly classified are redirected by the CEU to the HPU for re-processing. In this manner, the HPU soleley needs to process a fraction of the system workload reducing its memory and compute requirements. Moreover, the accuracy loss that is induced due to the extreme model quantisation of the LPU is restored to meet the user-specified error threshold.

%At run time, input samples are initially processed in batches by the LPU. The CEU examines the outputs of the LPU and determines the classification confidence for each sample. The samples that are detected as wrongly classified are redirected by the CEU to the HPU for re-processing. In this manner, the accuracy loss that is induced due to the extreme model quantisation of the LPU is restored to meet the user-specified error threshold.

 \subsection{Quantisation}
  		    \vspace{-0.05cm}
 Arithmetic precision reduction is a widely studied technique which exploits the inherent redundancy of CNNs to considerably reduce the memory bandwidth and footprint requirements, minimise power consumption and achieve higher performance. 
 %
 %Model size reduction is widely studied in CNNs to exploit the inherent redundancy that is present in most commonly studied networks. Specifically network quantisation allows the low-precision, fixed-point representation of CNN parameters which can significantly reduce memory bandwidth and footprint requirements, minimise the power consumption and lead to higher performance. Even if employing fine tuning of the model results to larger model compression or achieves higher accuracy results, it requires access to the full training set and framework; which in most cases does not form a valid assumption.
 %
 \textit{CascadeCNN} employs a fine-grained search space across possible precision quantisation schemes, that allows determining the number of integer and fractional bits of weight and activation values by introducing a different scaling factor for each layer. In this dynamic fixed-point approach, the wordlength is kept uniform across layers with a different scaling factor for each layer.
 %However, the wordlength is selected to be uniform across layers in order to derive a single hardware architecture for the whole network. 
 For each explored wordlength, statistics regarding the quantisation effect of each layer on the application-level accuracy are extracted using the user-provided evaluation set. The per-layer statistics are used to guide the exploration to the combination of scaling factors that achieve the highest accuracy for each explored wordlength. In contrast to other frameworks, \textit{CascadeCNN} selects for the LPU a precision that achieves intermediate application-level accuracy, but with significantly higher performance when mapped on its custom precision-optimised hardware units. All input samples are processed by the LPU to obtain a rapid classification decision, which is then fed to the Confidence Evaluation Unit. A wordlength that achieves an accuracy that complies with the user-specified error margins is selected for the HPU. 
 
 Since the reduced-precision model employed by the LPU is derived by straight quantisation (without retraining), its parameters are extracted at run time in hardware from the HPU's higher precision model. As a result of this weight-sharing approach, the memory footprint of the proposed cascade system remains the same as in the case of a single-stage architecture employing the HPU's model.

   		    \vspace{-0.23cm}
\subsection{Confidence Evaluation}
  		    \vspace{-0.12cm}
The \textit{CascadeCNN} tool allows the exploration of extreme quantisation schemes for the LPU, by aiming to identify potentially misclassified inputs based on the confidence of the LPU classification prediction. %Having induced an application-level accuracy loss due to the extreme quantisation scheme that is employed by \textit{CascadeCNN}, all classification predictions are evaluated based on their confidence to identify potentially misclassified samples... 
To estimate %the confidence of a classification prediction,
this confidence, we build on the work of \cite{joshi2009multi} by generalising the proposed Best-vs-Second-Best (BvSB) metric, which was previously examining solely binary classification problems. Our generalised BvSB (gBvSB) metric is described as:
%  Having suffered an application level accuracy loss due to the extreme quantisation scheme that is employed by \textit{CascadeCNN}, all classification predictions are evaluated based on their confidence to identify potentially misclassified samples that need to be processed by the higher precision unit which is chosen to meet the user-specified error thresholds. To identify the confidence of a classification prediction, we built on the work of [], by generalising the proposed BvSB metric, which was previously examining binary classification problems. Our generalised Best-vs-Second-Best metric is described as:
    \vspace{-0.18cm}
\begin{equation}
    \vspace{-0.16cm}
  \text{gBvSB}_{<M,N>}(\boldsymbol{p}) = \sum_{i=1}^{M}p_i - \sum_{j=M+1}^{N}p_j
\end{equation}
where $p_i$ denotes the i-th element of the sorted probability vector $\boldsymbol{p}$ of the prediction and $M$ and $N$ are tunable parameters of gBvSB. In this context, a prediction is considered confident, and thus the processing ends on the low-precision unit, when \mbox{$\text{gBvSB}_{<M,N>}(\boldsymbol{p}) \ge th$}
%\begin{equation}
%  \text{gBvSB}_{<M,N>}(\boldsymbol{p}) \ge th
%\end{equation}
where $M$, $N$ and threshold $th$ form tunable parameters whose values are automatically determined % after an exhaustive search over 
using the evaluation set data and the user-specified error tolerance. In this manner, the degree of uncertainty on the classification decision is based on how spiky the sorted probability distribution of the CNN's prediction is. %This metric is determining the degree of uncertainty on the classification decision based on how spiky the sorted probability distribution of CNN's prediction is. It should be noted that a trade-off exists between the number of samples that were correctly classified by the low-precision unit and are forwarded for re-processing to the high-precision unit, and the number of misclassified samples that the evaluator failed to identify.

% where $p_i$ denoted the i-th element of the sorted probability vector of the prediction and $M$, $N$ and threshold $th$ form tunable parameters whose values are automatically determined after an exhaustive search over the evaluation set data, based on the user-specified error tolerance. This metric is determining the degree of uncertainty on the classification decision based on how spiky the sorted probability distribution of CNN's prediction is. It should be noted that a trade-off exists between the number of samples that were correctly classified by the low-precision unit and are forwarded for re-processing to the high-precision unit, and the number of misclassified samples that the evaluator failed to identify. 
 
   		    \vspace{-0.23cm}
 \subsection{Architecture}
   		    \vspace{-0.12cm}
%  A scalable, fine-grained hardware architecture is designed to take advantage of the available resources of a given FPGA device and exploit higher degrees of parallelism as the wordlength of parameter representation drops. This architecture is related to the work of [], who proposed mapping all convolution operations to matrix multiplication. Similarly, batched FC layers can be mapped to the same matrix-multiplication unit, implemented at a tilled manner, across all matrix dimensions, on a tailored hardware architecture deployed to the FPGA device. MACC units implemented both on LUTs and DSPs, are group into Processing Elements that perform dot-product operations. Performance modelling based on roofline model [], and design space exploration are employed to determine the optimal tile-size parameters that trade between different levels of parallelism (number of PEs and MACCs-per-PE) for a given CNN,FPGA pair for each particular wordlength, achieving the highest throughput across different layers. The batch of inputs computed by each stage before reconfiguration of the device, is also processed on a tilled manner to achieve better balance between the workload(?) of typical Convolutional and FC layers. 

A scalable, fine-grained hardware architecture is designed that is able to execute CNN inference, scale its performance with the resources of a target FPGA and exploit higher degrees of parallelism as the wordlength of activation and weight representation decreases. The core of the architecture is a matrix multiplication (\textit{MM}) unit, parametrised with respect to the tiling of each matrix dimension and the arithmetic precision of both activations and weights. The \textit{MM} unit comprises Multiply-Accumulate (MACC) units, grouped into Processing Elements (PEs) that perform dot-product operations (shown in Fig. \ref{fig:arch}). By casting convolution operations as matrix multiplications and using batch processing for fully-connected (FC) layers, both CONV and FC layers are mapped on the \textit{MM} unit. 

%In contrast with the majority of the existing FPGA-based CNN accelerators, the proposed architecture utilises both LUTs and DSPs to implement its MACC units. With smaller wordlengths being less LUT-costly, employing LUT-based arithmetic units alongside DSP-based units enables the proposed architecture to reach higher performance by instantiating a higher number of MACC units as the wordlength decreases.

Given a CNN-FPGA pair and a particular wordlength, \textit{CascadeCNN} searches the architectural design space by means of a roofline-based performance model \cite{williams2009roofline} in order to determine the highest performing configuration of the architecture. The configurable parameters comprise the matrix tile sizes, that correspond to different levels of parallelism in terms of number of PEs and MACCs-per-PE. In this manner, \textit{CascadeCNN} generates two architectures, the LPU and the HPU, which are optimised for different wordlengths.

 \vspace{-0.2cm}
\section{Evaluation}
 \vspace{-0.1cm}
To evaluate the proposed toolflow, we target image classification using pretrained models on the ImageNet \cite{deng2009imagenet} dataset. \textit{CascadeCNN} is provided with models of VGG-16 \cite{simonyan2014very} and AlexNet \cite{krizhevsky2012imagenet}, along with a small subset of the ImageNet validation set as an evaluation set (200 labelled samples), targeting two different FPGA platforms, Xilinx Zynq ZC706 and UltraScale+ ZCU102. %Experiments are conducted on a wide range of error-tolerance. Matlab 2017a is used to investigate the quantised fixed-point network behaviour and determine the highest achieving quantisation scheme for each wordlength, as well as to obtaining network predictions on the evaluation set to tune the CEU parameters. The custom hardware architecture is synthesised and placed-and-routed using the Xilinx Vivado HLS and Vivado Design Suite (v17.2) and evaluated on the Xilinx Zynq ZC706 and UltraScale+ ZCU102 boards. 
%For both VGG-16 and AlexNet, \textit{CascadeCNN} yields a wordlength of 4 bits for the LPU. As shown on the left y-axis of Fig. \ref{fig:prec}, the selected 4-bit quantisation scheme introduces a 14.38\% and 18.65\% degradation in classification accuracy compared to an 8-bit precision respectively. Based on the attainable performance of the design points chosen by design space exploration (also shown on the right y-axis in Fig. \ref{fig:prec}), the LPU 4-bit architectures achieve higher throughput by 2.14$\times$ on VGG-16 and 2.08$\times$ on AlexNet. 

\begin{figure}%[h]  
	\centering
	\includegraphics[trim =0mm 0mm 0mm 0mm, width=0.93\columnwidth]{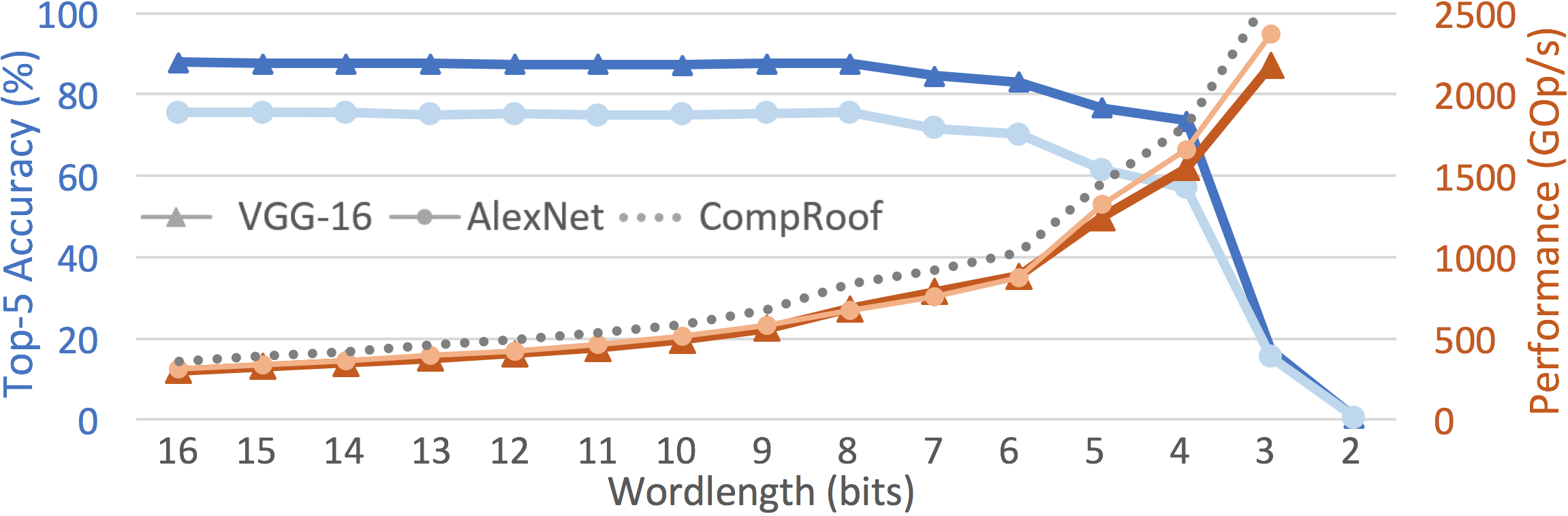}
		    \vspace{-0.4cm}
	\caption{ Top-5 classification accuracy on ImageNet and performance as a function of wordlength on Zynq ZC706. }
	\label{fig:prec}
		    \vspace{-0.25cm}
\end{figure}

\begin{figure}%[h]  
  			    \vspace{-0.10cm}
	\centering
	\includegraphics[trim =0mm 0mm 0mm 0mm, width=0.93\columnwidth]{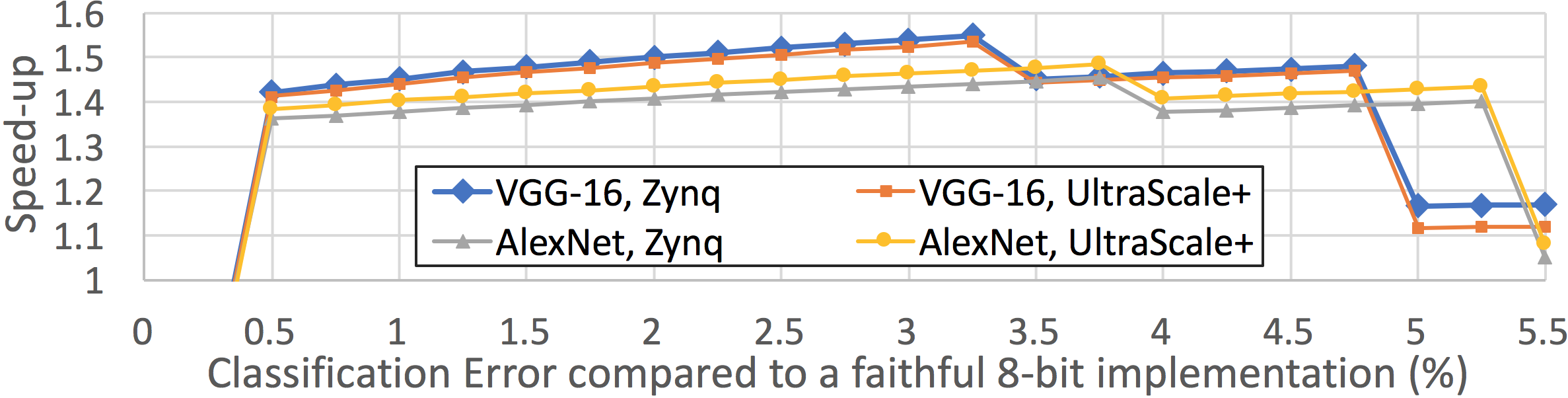}
			    \vspace{-0.4cm}
	\caption{\textit{CascadeCNN} speed-up }
	\label{fig:speedup}
			    \vspace{-0.60cm}
\end{figure}

For both VGG-16 and AlexNet, \textit{CascadeCNN} yields a wordlength of 4 bits for the LPU. The selected 4-bit quantisation scheme introduces a 14.38\% and 18.65\% degradation in classification accuracy compared to an 8-bit precision respectively (Fig. \ref{fig:prec}). The CEU parameters are tuned on the evaluation dataset to generate systems that introduce a wide range of classification errors, compared to a faithful 8-bit implementation. To evaluate the performance gains of \textit{CascadeCNN}, we compare the generated two-stage system for each error tolerance with a baseline single-stage architecture that is optimised with a quantisation scheme that achieves the same or better accuracy (ranging from 5 to 7 bit wordlengths). The achieved
%attainable 
speed-up on throughput is illustrated in Fig. \ref{fig:speedup} across a wide range of error thresholds. In the case of high error tolerance, the speed-up becomes less significant as the difference in wordlength between the LPU and the baseline design decreases. On both target platforms the performance has been improved by up to 55\% for VGG-16 and up to 48\% for AlexNet over the baseline design for the same resource budget and error tolerance. The proposed methodology can also be applied to other existing CNN accelerator architectures, with variable performance gains.

 \vspace{-0.2cm}
 \section{Conclusion}
This work presents \textit{CascadeCNN}, an automated toolflow for CNN inference acceleration exploiting the computation time-accuracy trade-off. %to boost the performance of quantised CNNs.
The cascaded two-stage architecture generated by the toolflow demonstrates a performance boost of up to 55\% for VGG-16 and 48\% for AlexNet compared to a single-stage baseline architecture for the same resource budget and error tolerance. 
  \vspace{-0.1cm}
{ \footnotesize %\small 
\section*{Acknowledgment}
 \vspace{-0.12cm}
The support of the EPSRC Centre for Doctoral Training in High Performance Embedded and Distributed Systems (HiPEDS, Grant Reference EP/L016796/1) is gratefully acknowledged. This work is also supported by EPSRC grant 1507723. 
 }
 
 \vspace{-0.15cm}
\bibliographystyle{plain}
\bibliography{main}

%\input{samplebody-conf}

%\bibliographystyle{ACM-Reference-Format}
%\bibliography{sample-bibliography}

\end{document}